\documentclass{article}

\usepackage{times}
\usepackage{graphicx} 
\usepackage{subfig}
\usepackage{amsmath,amsthm,amsfonts}
\usepackage{natbib}
\usepackage{float}

\let\origsection\section
\usepackage[section]{placeins}

\DeclareMathOperator*{\argmin}{arg\,min}

\newcommand{\ybar}{\bar{y}}

\usepackage{algorithm}
\usepackage{algorithmic}

\usepackage{hyperref}

\usepackage[accepted]{icml2016} 

\icmltitlerunning{Structured Prediction Energy Networks}

\begin{document} 

\twocolumn[
\icmltitle{Structured Prediction Energy Networks}

\icmlauthor{David Belanger}{belanger@cs.umass.edu}
\icmlauthor{Andrew McCallum}{mccallum@cs.umass.edu}
\icmladdress{College of Information and Computer Sciences, University of Massachusetts Amherst}

\icmlkeywords{structured prediction, deep learning}

\vskip 0.3in
]

\begin{abstract} 
We introduce \textit{structured prediction energy networks} (SPENs), a flexible framework for structured prediction.  A deep architecture is used to define an \textit{energy function} of candidate labels, and then predictions are produced by using back-propagation to iteratively optimize the energy with respect to the labels. This deep architecture captures dependencies between labels that would lead to intractable graphical models, and performs~\textit{structure learning} by automatically learning discriminative features of the structured output. One natural application of our technique is multi-label classification, which traditionally has required strict prior assumptions about the interactions between labels to ensure tractable learning and prediction. We are able to apply SPENs to multi-label problems with substantially larger label sets than previous applications of structured prediction, while modeling high-order interactions using minimal structural assumptions. Overall, deep learning provides remarkable tools for learning features of the inputs to a prediction problem, and this work extends these techniques to learning features of structured outputs. Our experiments provide impressive performance on a variety of benchmark multi-label classification tasks, demonstrate that our technique can be used to provide interpretable structure learning, and illuminate fundamental trade-offs between feed-forward and iterative structured prediction. 
\end{abstract} 

\origsection{Introduction}
\label{sec:intro}
Structured prediction is an important problem in a variety of machine learning domains. Consider an input $x$ and structured output $y$, such as a labeling of time steps, a collection of attributes for an image, a parse of a sentence, or a segmentation of an image into objects. Such problems are challenging because the number of candidate $y$ is exponential in the number of output variables that comprise it. As a result, practitioners encounter \textit{computational} considerations, since prediction requires searching an enormous space, and also \textit{statistical} considerations, since learning accurate models from limited data requires
reasoning about commonalities between distinct structured outputs. Therefore, structured prediction is fundamentally a problem of representation, where the representation must capture both the discriminative interactions between $x$ and $y$ and also allow for efficient combinatorial optimization over $y$. With this perspective, it is unsurprising that there are natural combinations of structured prediction and deep learning, a powerful framework for representation learning. 

We consider two principal approaches to structured prediction: (a) as a \textit{feed-forward} function $y = f(x)$, and (b) using an \textit{energy-based} viewpoint $y = \arg \min_{y'} E_x(y')$~\citep{lecun2006tutorial}. Feed-forward approaches include, for example, predictors using local convolutions plus a classification layer~\citep{collobert2011natural}, fully-convolutional networks ~\citep{long_shelhamer_fcn}, or sequence-to-sequence predictors~\citep{sutskever2014sequence}. Here, end-to-end learning can be performed easily using gradient descent. In contrast, the energy-based approach may involve non-trivial optimization to perform predictions, and includes, for example, conditional random fields (CRFs)~\citep{lafferty2001conditional}. From a modeling perspective, energy-based approaches are desirable because directly parametrizing $E_x(\cdot)$ provides practitioners with better opportunities to utilize domain knowledge about properties of the structured output. Furthermore, such a parametrization may be more parsimonious, resulting in improved generalization from limited data. On the other hand, prediction and learning are more complex. 

For energy-based prediction, prior applications of deep learning have mostly followed a two-step construction: first, choose an existing model structure for which the search problem $y = \arg \min_{y'} E_x(y')$ can be performed efficiently, and then express the dependence of $E_x(\cdot)$ on $x$ via a deep architecture. For example, the tables of potentials of an undirected graphical model can be parametrized via a deep network applied to $x$~\citep{lecun2006tutorial,collobert2011natural,DBLP:journals/corr/JaderbergSVZ14b,DBLP:journals/corr/HuangXY15,DBLP:journals/corr/SchwingU15,ChenSchwingICML2015}. The advantage of this approach is that it employs deep architectures to perform representation learning on $x$, while leveraging existing algorithms for combinatorial prediction, since the dependence of $E_x(y')$ on $y'$ remains unchanged. In some of these examples, exact prediction is intractable, such as for loopy graphical models,  and standard techniques for learning with approximate inference are employed. An alternative line of work has directly maximized the performance of iterative approximate prediction algorithms by performing back-propagation through the iterative procedure~\citep{stoyanov2011empirical,domke2013learning,hershey2014deep,crfasrnn_iccv2015}. 

All of these families of deep structured prediction techniques assume a particular graphical model for $E_x(\cdot)$ a-priori, but this construction perhaps imposes an excessively strict inductive bias. Namely, practitioners are unable to use the deep architecture to perform \textit{structure learning}, representation learning that discovers the interaction between different parts of $y$. In response, this paper explores \textit{Structured Prediction Energy Networks} (SPENs), where a deep architecture encodes the dependence of the energy on $y$, and predictions are obtained by approximately minimizing the energy iteratively via gradient descent. 

Using gradient-based methods to predict structured outputs was mentioned in~\citet{lecun2006tutorial}, but applications have been limited since then. Mostly, the approach has been applied for alternative goals, such as generating adversarial examples ~\citep{42503,goodfellow2014explaining}, embedding examples in low-dimensional spaces~\citep{le2014distributed}, or image synthesis~\citep{deepdream,DBLP:journals/corr/GatysEB15a,texture2015}. This paper provides a concrete extension of the ‘implicit regression’ approach of~\citet{lecun2006tutorial} to structured objects, with a target application (multi-label classification), a family of candidate architectures (Section~\ref{sec:arch}), and a training algorithm (a structured SVM~\citep{koller2004max,tsochantaridis2004support}). 

Overall, SPENs offer substantially different tradeoffs than prior applications of deep learning to structured prediction. Most energy-based approaches form predictions using optimization algorithms that are tailored to the problem structure, such as message passing for loopy graphical models. Since SPEN prediction employs gradient descent, an extremely generic algorithm, practitioners can explore a wider variety of differentiable energy functions. In particular, SPENS are able to model high-arity interactions that would result in unmanageable treewidth if the problem was posed as an undirected graphical model. On the other hand, SPEN prediction lacks algorithmic guarantees, since it only performs local optimization of the energy. 

SPENs are particularly well suited to multi-label classification problems. These are naturally posed as structured prediction, since the labels exhibit rich interaction structure. However, unlike problems with grid structure, where there is a natural topology for prediction variables, the interactions between labels must be learned from data. Prior applications of structured prediction, eg. using CRFs, have been limited to small-scale problems, since the techniques' complexity, both in terms of the number of parameters to estimate and the per-iteration cost of algorithms like belief propagation, grows at least quadratically in the number of labels $L$~\citep{ghamrawi2005collective,finley2008training,meshi2010learning,petterson2011submodular}. For SPENs, though, both the per-iteration prediction complexity and the number of parameters scale linearly in $L$. We only impose mild prior assumptions about labels' interactions: they can be encoded by a deep architecture. Motivated by recent compressed sensing approaches to multi-label classification~\citep{hsu2009multi,kapoor2012multilabel}, we further assume that the first layer of the network performs a small set of linear projections of the prediction variables. This provides a particularly parsimonious representation of the energy function and an interpretable tool for structure learning. 

On a selection of benchmark multi-label classification tasks, the expressivity of our deep energy function provides accuracy improvements against a variety of competitive baselines, including a novel adaptation of the `CRF as RNN' approach of~\citet{crfasrnn_iccv2015}. We also offer experiments contrasting SPEN learning with alternative SSVM-based techniques and analyzing the convergence behavior and speed-accuracy tradeoffs of SPEN prediction in practice. Finally, experiments on synthetic data with rigid mutual exclusivity constraints between labels demonstrate the power of SPENs to perform structure learning and illuminate important tradeoffs in the expressivity and parsimony of SPENs vs. feed-forward predictors. We encourage further application of SPENs in various domains.

\section{Structured Prediction Energy Networks}
\label{sec:dmfn}

For many structured prediction problems, an $x \rightarrow y$ mapping can be defined by posing $y$ as the solution to a potentially non-linear combinatorial optimization problem, with parameters dependent on $x$~\citep{lecun2006tutorial}:
\begin{equation}
\begin{aligned}
\underset{y}{\text{min}} & \quad \mathrm{E_x}(y) && \text{subject to \quad} &  y \in \{0,1\}^L.
\end{aligned} \label{eq:comb}
\end{equation}
This includes binary CRFs, where there is a coordinate of $y$ for every node in the graphical model.
Problem~\eqref{eq:comb} could be rendered tractable by assuming certain structure (e.g., a tree) for the energy function $E_x(\cdot)$. Instead, we consider general $E_x(\cdot)$, but optimize over a convex relaxation of the constraint set:
\begin{align}\label{eq:relax}
\underset{\bar{y}}{\text{min}} & \quad \mathrm{E_x}(\bar{y}) && \text{subject to \quad} &  \bar{y} \in [0,1]^L.
\end{align} 
 In general, $E_x(\ybar)$ may be non-convex, where exactly solving~\eqref{eq:relax} is intractable. A reasonable approximate optimization procedure, however, is to minimize~\eqref{eq:relax} via gradient descent, obtaining a local minimum.  Optimization over $[0,1]^L$ can be performed using projected gradient descent, or entropic mirror descent by normalizing over each coordinate~\citep{beck2003mirror}.  We use the latter because it maintains iterates in $(0,1)^L$, which allows using energy functions and loss functions that diverge at the boundary. 

There are no guarantees that our predicted $\ybar$ is nearly 0-1. In some applications, we may round $\ybar$ to obtain predictions that are usable downstream. It may also be useful to maintain `soft' predictions, eg. for detection problems.

In the posterior inference literature, \textit{mean-field} approaches also consider a relaxation from $y$ to $\ybar$, where $\ybar_i$ would be interpreted as the marginal probability that $y_i = 1$~\citep{jordan1999introduction}.  Here, the practitioner starts with a probabilistic model for which inference is intractable, and obtains a mean-field objective when seeking to perform approximate variational inference. We make no such probabilistic assumptions, however, and instead adopt a discriminative approach by directly parametrizing the objective that the inference procedure optimizes. 

Continuous optimization over $\ybar$ can be performed using black-box access to a gradient subroutine for $E_x(\ybar)$. Therefore,  it is natural to parametrize $E_x(\ybar)$ using deep architectures, a flexible family of multivariate function approximators that provide efficient gradient calculation. 

A SPEN parameterizes $E_x(\ybar)$ as a neural network that takes both $x$ and $\ybar$ as inputs and returns the energy (a single number). In general, a SPEN consists of two deep architectures. 
First, the \textit{feature network} F($x$) produces an $f$-dimensional feature representation for the input.  
Next, the energy $E_x(\ybar)$ is given by the output of the \textit{energy network} $E(F(x),\ybar)$. Here, $F$ and $E$ are arbitrary deep networks. 

Note that the energy only depends on $x$ via the value of $F(x)$. During iterative prediction, we improve efficiency by precomputing $F(x)$ and not back-propagating through $F$ when differentiating the energy with respect to $\ybar$.

\section{Example SPEN Architecture}
\label{sec:arch}

We now provide a more concrete example of the architecture for a SPEN. All of our experiments use the general configuration described in this section. We denote matrices in upper case and vectors in lower case. We use  $g()$ to denote a coordinate-wise non-linearity function, and may use different non-linearities, eg. sigmoid vs. rectifier, in different places. Appendix~\ref{app:architecture-figure} provides a computation graph for this architecture. 

For our feature network,  we employ a simple 2-layer network:
\begin{equation}
F(x) = g(A_2 g(A_1 x)). \label{eq:feats}
\end{equation}
Our energy network is the sum of two terms. First, the \textit{local energy network} scores $\ybar$ as the sum of $L$ linear models:
\begin{equation}
E^{\text{local}}_x(\ybar) = \sum_{i = 1}^L \ybar_i b_i^\top F(x). \label{eq:local}
\end{equation}
Here, each $b_i$ is an $f$ dimensional vector of parameters for each label. 

This score is added to the output of the \textit{global energy network}, which scores configurations of $\ybar$ independent of $x$:
\begin{equation}
E^{\text{label}}_x(\ybar) = c_2 ^\top g(C_1 \ybar). \label{eq:le}
\end{equation}
The product $C_1 \ybar$ is a  set of learned affine (linear + bias) measurements of the output,  that capture salient features of the labels used to model their dependencies. By learning the \textit{measurement matrix} $C_1$ from data, the practitioner imposes minimal assumptions a-priori on the interaction structure between the labels, but can model sophisticated interactions by feeding $C_1 \ybar$ through a non-linear function. This has the additional benefit that the number of parameters to estimate grows linearly in $L$. In Section~\ref{sec:structure-learning}, we present experiments exploring the usefulness of the measurement matrix as a means to perform structure learning. 

In general, there is a tradeoff between using increasingly expressive energy networks and being more vulnerable to overfitting. In some of our experiments, we add another layer of depth to~\eqref{eq:le}. It is also natural to use a global energy network that conditions on $x$, such as:
\begin{equation}
E^{\text{cond}}_x(\ybar) = d_2 ^\top g(D_1 [\ybar; F(x)]). \label{eq:cle}
\end{equation}
Our experiments consider tasks with limited training data, and here we found that using a data-independent energy~\eqref{eq:le} helped prevent overfitting, however. 

Finally, it may be desirable to choose an architecture for $g$ that results in a convex problem in $y$~\citet{amosinput2016}. Our experiments select $g$ based on accuracy, rather than algorithmic guarantees resulting from convexity. 

\subsection{Conditional Random Fields as SPENs}
There are important parallels between the example SPEN architecture given above and the parametrization of a CRF~\citep{lafferty2001conditional}. Here, we use CRF to refer to any structured linear model, which may or may not be trained to maximize the conditional log likelihood. For the sake of notational simplicity, consider a fully-connected pairwise CRF with local potentials that depend on $x$, but data-independent pairwise potentials. Suppose we apply $E_x(\cdot)$ directly to $y$, rather than to the relaxation $\ybar$. The corresponding global energy net would be:
\begin{equation}
E^{\text{crf}}_x(y) =  y^\top S_1 y + s^\top y. \label{eq:crf}
\end{equation}

In applications with large label spaces,~\eqref{eq:crf} is troublesome in terms of both the statistical efficiency of parameter estimation and the computational efficiency of prediction because of the quadratic dependence on $L$. Statistical issues can be mitigated by imposing parameter tying of the CRF potentials, using a low-rank assumption, eg. \citep{srikumar2014learning,DBLP:conf/icml/JerniteRS15}, or using a deep architecture to map $x$ to a table of CRF potentials~\citep{lecun2006tutorial}.  Computational concerns can be mitigated by choosing a sparse graph. This is difficult for practitioners when they do not know the dependencies between labels a-priori. Furthermore, modeling high-order interactions than pairwise relationships is very expensive with a CRF, but presents no extra cost for SPENs. 
Finally, note that using affine, rather than linear, measurements in $C_1$ is critical to allow SPENs to be sufficiently universal to model dissociativity between labels, a common characteristic of CRFs. 

For CRFs, the interplay between the graph structure and the set of representable conditional distributions is well-understood~\citep{koller2009probabilistic}. However, characterizing the representational capacity of SPENs is more complex, as it depends on the general representational capacity of the deep architecture chosen.

\section{Learning SPENs}

In Section~\ref{sec:dmfn}, we described a technique for producing predictions by performing continuous optimization in the space of outputs. Now we discuss a gradient-based technique for learning the parameters of the network $E_x(\ybar)$. 

In many structured prediction applications, the practitioner is able to interact with the model in only two ways: (1) evaluate the model's energy on a given value of $y$, and (2) minimize the energy with respect to the $y$.  This occurs, for example, when predicting combinatorial structures such as bipartite matchings and graph cuts.  A  popular technique in these settings is the structured support vector machine (SSVM)~\citep{koller2004max,tsochantaridis2004support}.

If we assume (incorrectly) that our prediction procedure is not subject to optimization errors, then (1) and (2) apply to our model and it is straightforward to train using an SSVM. This ignores errors resulting from the potential non-convexity of $E_x(\ybar)$ or the relaxation from $y$ to $\ybar$. However, such an assumption is a reasonable way to construct an approximate learning procedure.

Define $\Delta(y_p,y_g)$ to be an error function between a prediction $y_p$ and the ground truth $y_g$, such as the Hamming loss.  Let $\left[\cdot\right]_+ = \max(0,\cdot)$. The SSVM minimizes:
\begin{equation}
\sum_{\{x_i,y_i\}} \max_{y} \left[\Delta(y_i,y) - E_{x_i}(y) + E_{x_i}(y_i)\right]_+. \label{eq:ssvm}
\end{equation}
Here, the $\left[\cdot\right]_+$ function is redundant when performing exact energy minimization. We require it, however, because gradient descent only performs approximate minimization of the non-convex energy. Note that the signs in~\eqref{eq:ssvm} differ from convention because here prediction minimizes $E_x(\cdot)$. We minimize our loss with respect to the parameters of the deep architecture $E_x$ using mini-batch stochastic gradient descent. For $\{x_i,y_i\}$, computing the subgradient of~\ref{eq:ssvm} with respect to the prediction requires \textit{loss-augmented inference}:
\begin{equation}
y_p = \argmin_{y} \left(-\Delta(y_i,y) + E_{x_i}(y)\right). \label{eq:loss-aug}
\end{equation}
With this, the subgradient of~\eqref{eq:ssvm} with respect to the model parameters is obtained by back-propagation through $E_x$. 

We perform loss-augmented inference by again using gradient descent on the relaxation $\ybar$, rather than performing combinatorial optimization over $y$. 
Since $\Delta$ is a discrete function such as the Hamming loss, we need to approximate it with a differentiable surrogate loss, such as the squared loss or log loss. For the log loss, which diverges at the boundary, mirror descent is crucial, since it maintains $\ybar \in (0,1)^L$. The objective~\eqref{eq:ssvm} only considers the energy values of the ground truth and the prediction, ensuring that they're separated by a margin, not the actual ground truth and predicted labels~\eqref{eq:loss-aug}. Therefore, we do not round the output of~\eqref{eq:loss-aug} in order to approximate a  subgradient of ~\eqref{eq:ssvm}; instead, we use the $\ybar$ obtained by approximately minimizing~\eqref{eq:loss-aug}. 

Training undirected graphical models using an SSVM loss is conceptually more attractive than training SPENs, though. In loopy graphical models, it is tractable to solve the LP relaxation of MAP inference using graph-cuts or message passing techniques, eg. \citep{boykov2004experimental,globerson2008fixing}. Using the LP relaxation, instead of exact MAP inference, in the inner loop of CRF SSVM learning is fairly benign, since it is guaranteed to over-generate margin violations in~\eqref{eq:ssvm}~\citep{kulesza2007structured,finley2008training}. On the other hand, SPENs may be safer from the perils of in-exact optimization during learning than training CRFs with the log loss.  As~\citet{lecun2005loss} discuss, the loss function for un-normalized models is unaffected by ``low-energy areas that are never reached by the inference algorithm ."
 
Finally, we have found that it is useful to initialize the parameters of the feature network by first training them using a simple local classification loss, ignoring any interactions between coordinates of $y$. For problems with very limited training data, we have found that overfitting can be lessened by keeping the feature network's parameters fixed when training the global energy network parameters.

\section{Applications of SPENs}
SPENs are a natural model for multi-label classification, an important task in a variety of machine learning applications. The data consist of $\{x,y\}$ pairs, where $ y = \{y_1, \ldots, y_L\} \in \{0,1\}^L$ is a set of multiple binary labels we seek to predict and $x$ is a feature vector. In many cases, we are given no structure among the $L$ labels a-priori, though the labels may be quite correlated. SPENs are a very natural model for multi-label classification because learning the measurement matrix $C_1$ in~\eqref{eq:le} provides an automatic method for discovering this interaction structure. Section~\ref{sec:rel-mlc} discusses related prior work. 

SPENs are very general, though, and can be applied to any prediction problem that can be posed, for example, as MAP inference in an undirected graphical model. In many applications of graphical models, the practitioner employs certain prior knowledge about dependencies in the data to choose the graph structure, and certain invariances in the data to impose parameter tying schemes. For example, when tagging sequences with a linear-chain CRF, the parameterization of local and pairwise potential functions is shared across time. Similarly, when applying a SPEN, we can express the global energy net~\eqref{eq:le} using temporal convolutions, ie. $C_1$ has a repeated block-diagonal structure. Section~\ref{sec:details} describes details for improving the accuracy and efficiency of SPENs in practice. 

\section{Related Work}

\subsection{Multi-Label Classification}
\label{sec:rel-mlc}

The most simple multi-label classification approach is to independently predict each label $y_i$ using a separate classifier, also known as the `binary relevance model'. This can perform poorly, particularly when certain labels are rare or some are highly correlated. Modeling improvements use max-margin or ranking losses that directly address the multi-label structure~\citep{elisseeff2001kernel,godbole2004discriminative,bucak2009efficient}. 

Correlations between labels can be modeled explicitly using models with low-dimensional embeddings of labels~\citep{ji2009linear,cabral2011matrix,yu2014largescal,DBLP:journals/corr/BhatiaJK0V15}. This can be achieved, for example, by using low-rank parameter matrices. In the SPEN framework, such a model would consist of a linear feature network~\eqref{eq:feats} of the form $F(x) = A_1 x$, where $A_1$ has fewer rows than there are target labels, and no global energy network. While the prediction cost of such methods grows linearly with $L$, these models have limited expressivity, and can not capture strict structural constraints among labels, such as mutual exclusivity and implicature. By using a non-linear multi-layer perceptron (MLP) for the feature network with hidden layers of lower dimensionality than the input, we are able to capture similar low-dimensional structure, but also capture interactions between outputs. In our experiments, is a MLP competitive baseline that has been under-explored in prior work. 

It is natural to approach multi-label classification using structured prediction, which models interactions between prediction labels directly. However, the number of parameters to estimate and the per-iteration computational complexity of these models grows super-linearly in $L$~\citep{ghamrawi2005collective,finley2008training,meshi2010learning,petterson2011submodular} , or requires strict assumptions about labels' depencies~\citep{read2011classifier,ews2}. 

Our parametrization of the global energy network~\eqref{eq:le} in terms of linear measurements of the labels is inspired by prior approaches using compressed sensing and error-correcting codes for multi-label classification~\citep{hsu2009multi,hariharan2010large,kapoor2012multilabel}. However, these rely on assumptions about the sparsity of the true labels or prior knowledge about label interactions, and often do not learn the measurement matrix from data. We do not assume that the labels are sparse. Instead, we assume their interaction can be parametrized by a deep network applied to a set of linear measurements of the labels.

\subsection{Deep Structured Models}
\label{sec:dsm}
It is natural to parametrize the potentials of a CRF using deep features~\citep{lecun2006tutorial,collobert2011natural,DBLP:journals/corr/JaderbergSVZ14b,DBLP:journals/corr/HuangXY15,Chen2014,DBLP:journals/corr/SchwingU15,ChenSchwingICML2015}. 
Alternatively, non-iterative feed-forward predictors can be constructed using structured models as motivation~\citep{stoyanov2011empirical,domke2013learning,kunisch2013bilevel,hershey2014deep,li2014,crfasrnn_iccv2015}. Here, a  model family is chosen, along with an iterative approximate inference technique for the model. The inference technique is then unrolled into a computation graph, for a fixed number of iterations, and parameters are learned end-to-end using backpropagation. This directly optimizes the performance of the approximate inference procedure, and is used as a baseline in our experiments.

While these techniques can yield expressive dependence on $x$ and improved training, the dependence of their expressivity and scalability on $y$ is limited, since they build on an underlying graphical model. They also require deriving model-structure-specific inference algorithms. 

\subsection{Iterative Prediction using Neural Networks}
Our use of backprogation to perform gradient-based prediction differs from most deep learning applications, where backpropagation is used to update the network parameters. However, backpropagation-based prediction has been useful in a variety of deep learning applications, including \textit{siamese networks}~\citep{bromley1993signature}, methods for generating adversarial examples~\citep{42503,goodfellow2014explaining}, methods for embedding documents as dense vectors~\citep{le2014distributed}, and successful techniques for image generation and texture synthesis~\citep{deepdream,DBLP:journals/corr/GatysEB15a,texture2015}. ~\citep{carreira2015human} propose an iterative structured prediction method for human pose estimation, where predictions are constructed incrementally as $y_{t+1} = y_t + \Delta(x,y_t)$. The $\Delta$ network is trained as a multi-variate regression task, by defining a ground truth trajectory for intermediate $y_t$.

\section{Experiments}
\subsection{Multi-Label Classification Benchmarks}
\label{sec:benchmarks}
\begin{table}[tb]
\begin{center}
\begin{tabular}{| c | c | c | c | c | c |}
\hline
  & BR & LR & MLP & DMF & SPEN  \\
\hline 
Bibtex  & 37.2 & 39.0  & 38.9 & 40.0 & \textbf{42.2}\\
\hline
Bookmarks & 30.7 &31.0 & 33.8 &  33.1 & \textbf{34.4}\\
\hline
Delicious & 26.5 &35.3 & \textbf{37.8} & 34.2 & 37.5\\
\hline
\end{tabular}
\caption{Comparison of various methods on 3 standard datasets in terms of F1 (larger is better). }
\label{tab:results}
\end{center}
\end{table}

Table~\ref{tab:results} compares SPENs to a variety of high-performing baselines on a selection of standard multi-label classification tasks. Dataset sizes, etc. are described in Table~\ref{tab:datasets}. We contrast SPENs with BR: independent per-label logistic regression; MLP: multi-layer perceptron with ReLU non-linearities trained with per-label logistic loss, ie. the feature network equation~\eqref{eq:feats} coupled with the local energy network equation~\eqref{eq:local}; and LR: the low-rank-weights method of~\citet{yu2014largescal}.  BR and LR results, are from~\citet{prlr:mmlnlp14}. The local energy of the SPEN is identical to the MLP.

We also compare to deep mean field (DMF), an instance of the deep `unrolling' technique described in Section~\ref{sec:dsm}. We consider 5 iterations of mean-field inference in a fully connected pairwise CRF with data-dependent pairwise factors, and perform end-to-end maximum likelihood training. Local potentials are identical to the MLP classifier, and their parameters are clamped to reduce overfitting (unlike any of the other methods, the DMF has $L^2$ parameters).  Details of our DMF predictor, which may be of independent interest, are provided in Section~\ref{sec:dmf-pred}. Note that we only obtained high performance by using pretrained unary potentials from the MLP. Without this, accuracy was about half that of Table~\ref{tab:results}. 

We report the example averaged (macro average) F1 measure. For Bibtex and Delicious, we tune hyperparameters by pooling the train and test data and sampling without replacement to make a split of the same size as the original. For Bookmarks, we use the same train-dev-test split as~\citet{prlr:mmlnlp14}.We seleced 15 linear measurements (rows of $C_1$ in~\eqref{eq:le}) for Bookmarks and Bibtex, and 5 for Delicious. Section~\ref{sec:hyper} describes additional choices of hyperparameters.  For SPENs, we obtain predictions by rounding $\bar{y}_i$ above a threshold tuned on held-out data. 

There are multiple key results in Table~\ref{tab:results}. First, SPENs are competitive compared to all of the other methods, including DMF, a structured prediction technique. While DMF scales computationally to moderate scales, since the algorithm in Section~\ref{sec:dmf-pred} is vectorized and can be run efficiently on a GPU, and can not scale statistically, since the pairwise potentials have so many parameters. As a result, we found it difficult to avoid overfitting with DMF on the Bookmarks and Delicious datasets. In fact, the best performance is obtained by using the MLP unary potentials and ignoring pairwise terms. Second, MLP, a technique that has not been treated as a baseline in recent literature, is surprisingly accurate as well. Finally, the MLP outperformed SPEN on the Delicious dataset. Here, we found that accurate prediction requires well-calibrated soft predictions to be combined with a confidence threshold. The MLP, which is trained with logistic regression, is better at predicting soft predictions than SPENs, which are trained with a margin loss. To obtain the SPEN result for Delicious in Table~\ref{tab:results}, we need to smooth the test-time prediction problem with extra entropy terms to obtain softer predictions.

Many multi-label classification methods approach learning as a missing data problem. Here, the training labels $y$ are assumed to be correct only when $y_i = 1$. When $y_i = 0$, they are treated as missing data, whose values can be imputed using assumptions about the rank~\citep{prlr:mmlnlp14} or sparsity~\citep{bucak2011multi,agrawal2013multi} of the matrix of training labels. For certain multi-label tasks, such modeling is useful because only positive labels are annotated. For example, the approach of~\citep{prlr:mmlnlp14} achieves 44.2 on the Bibtex dataset, outperforming our method, but only 33.3 on Delicious, substantially worse than the MLP or SPEN.  Missing data modeling is orthogonal to the modeling of SPENs, and we can combine missing data techniques with SPENs.

\subsection{Comparison to Alternative SSVM Approaches}
\label{sec:anal-ssvm}
\begin{table}[tb]
\begin{center}
\begin{tabular}{| c | c | c | c | c |}
 \hline 
EXACT  & LP & LBP & DMF & SPEN  \\
\hline 
20.2 $\pm$  .5   & 20.5 $\pm$ .5  & 24.3 $\pm$  .6 & 23 $\pm$ .2 & \textbf{20.0} $\pm$ .3\\
\hline 
\end{tabular}
\caption{Comparing different prediction methods, which are used both during SSVM training and at test time, using the setup of~\citet{finley2008training} on the Yeast dataset, with hamming error (smaller is better). SPENs perform comparably to EXACT and LP, which provide stronger guarantees for  SSVM training.}
\label{tab:results2} 
\end{center}
\end{table}

Due to scalability considerations, most prior applications of CRFs to multi-label classification have been restricted to substantially smaller $L$ than those considered in Table~\ref{tab:results}. In Table~\ref{tab:results2}, we consider the 14-label yeast dataset~\citep{elisseeff2001kernel}, which is the largest label space fit using a CRF in~\citet{finley2008training} and \citet{meshi2010learning}.  ~\citet{finley2008training} analyze the effects of inexact prediction on SSVM training and on test-time prediction. Table~\ref{tab:results2} considers exact prediction using an ILP solver, loopy belief propagation  (LBP), solving the LP relaxation, the deep mean-field network described in the previous section, and SPENs, where the same prediction technique is used at train and test time. All results, besides SPEN and DMF, are from~\citet{finley2008training}. The SPEN and DMF use linear feature networks. We report hamming error, using 10-fold cross validation. 

We use Table~\ref{tab:results2} to make two arguments. First, it provides justification for our use of the deep mean field network as an MRF baseline in the previous section, since it performs comparably to the other MRF methods in the table, and substantially better than LBP. Second, a key argument of~\citet{finley2008training} is that SSVM training is more effective when the train-time inference method will not under-generate margin violations. Here, LBP and SPEN, which both approximately minimize a non-convex inference objective, have such a vulnerability, whereas LP does not, since solving the LP relaxation provides a lower bound on the true solution to the value of ~\eqref{eq:loss-aug}. Since SPEN performs similarly to EXACT and LP, this suggests that perhaps the effect of inexact prediction is more benign for SPENs than for LBP. However, SPENs exhibit alternative expressive power to pairwise CRFs, and thus it is difficult to fully isolate the effect of SSVM training on accuracy.

\subsection{Structure Learning Using SPENs}
\label{sec:structure-learning}
Next, we perform experiments on synthetic data designed to demonstrate that the label measurement matrix, $C_1$ in the global energy network~\eqref{eq:le}, provides a useful tool for analyzing the structure of dependencies between labels. SPENs impose a particular inductive bias about the interaction between $x$ and $y$. Namely, the interactions between different labels in $y$ do not depend on $x$. Our experiments show that this parametrization allows SPENs to excel in regimes of limited training data, due to their superior parsimony compared to analogous feed-forward approaches. 

To generate data, we first draw a design matrix $X$ with 64 features, with each entry drawn from $N(0,1)$. Then, we generate a 64 x 16 weights matrix $A$, again from $N(0,1)$. Then, we construct $Z = X A$ and split the 16 columns of $Z$ into 4 consecutive blocks. For each block, we set $Y_{ij} = 1$ if $Z_{ij}$ is the maximum entry in its row-wise block, and 0 otherwise. We seek a model with predictions that reliably obey these within-block exclusivity constraints. 

\begin{figure}[tb]
\centering
\subfloat[ReLU]{
\includegraphics[width=0.4\columnwidth]{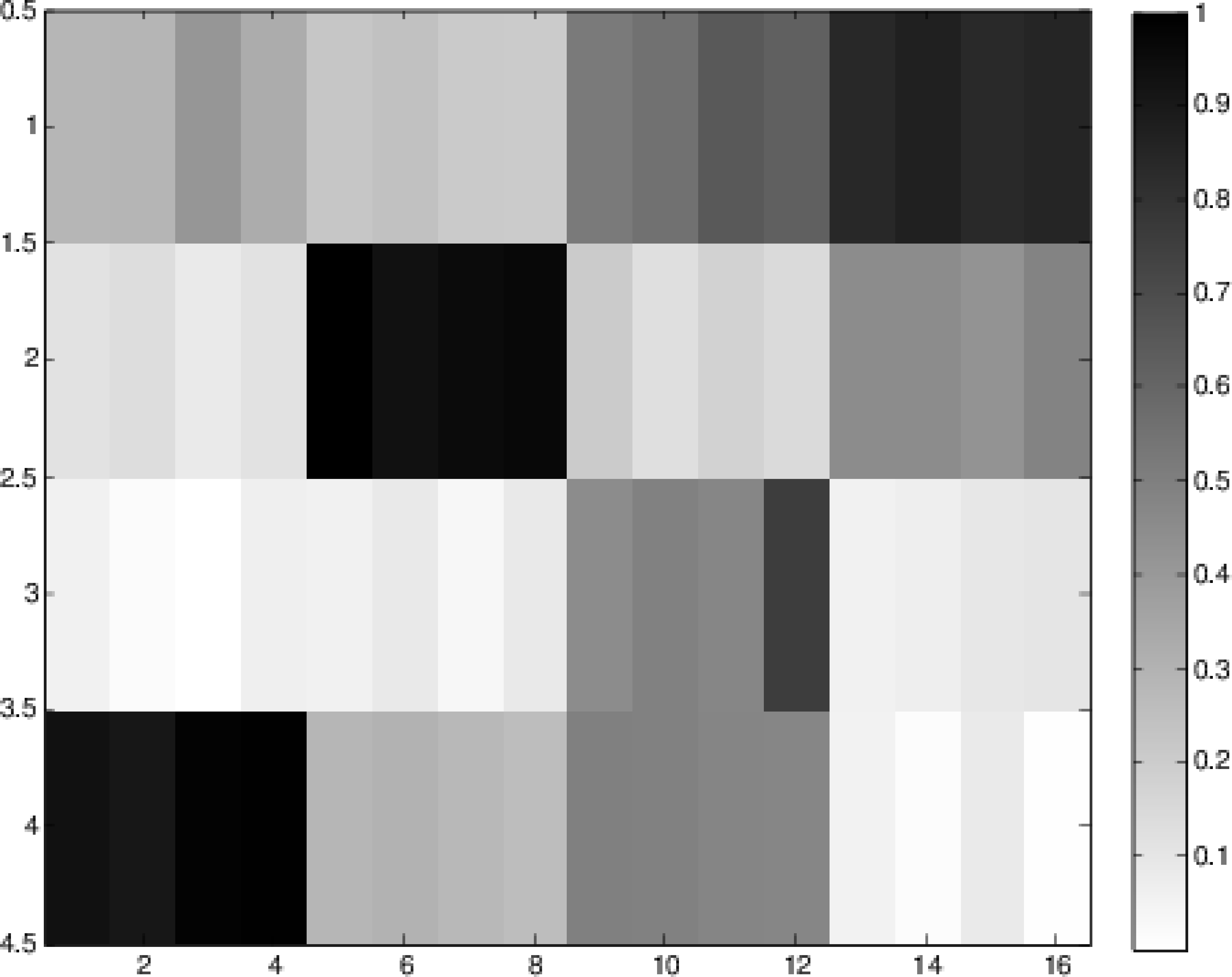}
}
\subfloat[HardTanh]{
\includegraphics[width=0.4\columnwidth]{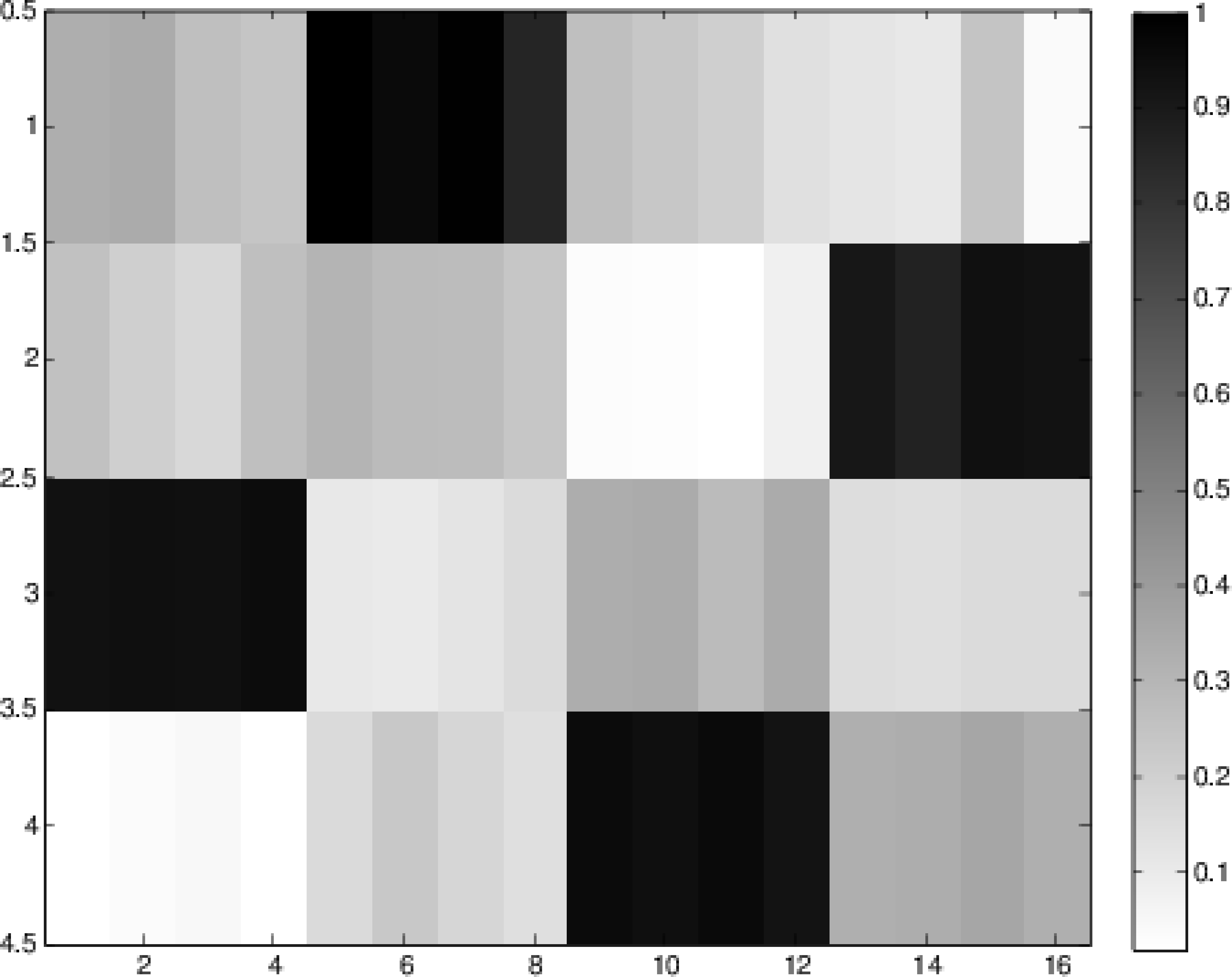}
}
\caption{Learned SPEN measurement matrices on synthetic data containing mutual exclusivity of labels within size-4 blocks, for two different choices of nonlinearity in the global energy network. 16 Labels on horizontal axis and 4 hidden units on vertical axis. }
\label{fig:structure}
\end{figure}

Figure~\ref{fig:structure} depicts block structure in the learned measurement matrix.  Measurements that place equal weight on every element in a block can be used to detect violations of the mutual exclusivity constraints characteristic of the data generating process. The choice of network architecture can significantly affect the interpretability of the measurement matrix, however. When using ReLU, which acts as the identity for positive activations, violations of the data constraints can be detected by taking linear combinations of the measurements (a), since multiple hidden units place large weight on some labels. This obfuscates our ability to perform structure learning by investigating the measurement matrix. On the other hand, since applying HardTanh to measurements saturates from above, the network learns to utilize each measurement individually, yielding substantially more interpretable structure learning in (b).
\begin{table}[tb]
\begin{center}
\begin{tabular}{| c | c |  c | c | c | }
 \hline 
\# train examples & Linear & 3-Layer MLP  &  SPEN\\
\hline 
1.5k & 80.0 & 81.6 & \textbf{91.5}  \\
\hline
15k & 81.8 & 96.3 & \textbf{96.7}  \\
\hline 
\end{tabular}
\caption{ Comparing performance (F1) on the synthetic task with block-structured mutual exclusivity between labels. Due to its parsimonious parametrization, the SPEN succeeds with limited data. With more data, the MLP performs comparably, suggesting that even rigid constraints among labels can be predicted in a feed-forward fashion using a sufficiently expressive architecture.}
\label{tab:resultsSynth} 
\end{center}
\end{table}

Next, in Table~\ref{tab:resultsSynth} we compare: a linear classifier, a 3-Layer ReLU MLP with hidden units of size 64 and 16, and a SPEN with a simple linear local energy network and a 2-layer global energy network with HardTanh activations and 4 hidden units. Using fewer hidden units in the MLP results in substantially poorer performance. We avoid using non-linear local energy network because we want to force the global energy network to capture all label interactions.

Note that the SPEN consistently outperforms the MLP, particularly when training on only 1.5k examples. In the limited data regime, their difference is because the MLP  has 5x more parameters, since we use a simple linear feature network in the SPEN. We also inject domain knowledge about the constraint structure when designing the global energy network's architecture.  Figure~\ref{fig:structure2} in the Appendix demonstrates that we can perform the same structure learning as in Figure~\ref{fig:structure} on this small training data. 

Next, observe that for 15k examples the performance of the MLP and SPEN are comparable. Initially, we hypothesized that the mutual exclusivity constraints of the labels could not be satisfied by a feed-forward predictor, and that reconciling their interactions would require an iterative procedure. However, it seems that a large, expressive MLP can learn an accurate predictor when presented with lots of examples.  Going forward, we would like to investigate the parsimony vs. expressivity tradeoffs of SPENs and MLPs.

\subsection{Convergence Behavior of SPEN Prediction}
 This section provides experiments analyzing the behavior of SPENs' test-time optimization in practice. All figures section appear in Appendix~\ref{app:convergence}. 

Prediction, both at train and test time, is performed in parallel in large minibatches on a GPU. Despite providing substantial speedups, this approach is subject to the `curse of the last reducer,' where unnecessary gradient computation is performed on instances for which optimization has already converged. Convergence is determined by relative changes in the optimization objective and absolute changes in the iterates’ values. In Figure~\ref{fig:minipage1} we provide a histogram of the iteration at which examples converge on the Bibtex dataset. The vast majority of examples converge (at around 20 steps) much before the slowest example (41 steps).  Predictions are often spiked at either 0 or 1, despite optimizing a non-convex energy over the set $[0,1]$. We expect that this results from the energy function being fit to 0-1 data. 

Since the speed of batch prediction is largely influenced by the worst-case iteration complexity, we seek ways to decrease this worst case while maintaining high aggregate accuracy. We hypothesize that prediction is slow on `hard' examples for which the model would have made incorrect predictions anyway. In response, we terminate prediction when a certain percentage of the examples have converged at a certain threshold. In Figure~\ref{fig:minipage2}, we vary this percentage from 50\% to 100\%, obtaining a 3-fold speedup at nearly no decrease in accuracy. In Figure~\ref{fig:minipage3}, we vary the tightness of the convergence threshold, terminating optimization when 90\% of examples have converged. This is also achieves a 3-fold speedup at little degradation in accuracy. Finally, Figures~\ref{fig:minipage1}-\ref{fig:minipage2} can be shifted by about 5 iterations to the left by initializing optimization at the output of the MLP. Section~\ref{sec:benchmarks} use configurations tuned for accuracy, not speed. 

Unsurprisingly, prediction using a feed-forward method, such as the above MLP or linear models, is substantially faster than a SPEN. The average total times to classify all 2515 items in the Bibtex test set are 0.0025 and 1.2 seconds for the MLP and SPEN, respectively. While SPENs are much slower, the speed per test example is still practical for various applications. Here, we used the termination criterion described above where prediction is halted when 90\% of examples have converged. If we were to somehow circumvent the curse-of-the-last-reducer issue, the average number of seconds of computation per example for SPENs would be 0.8 seconds.  Note that the feature network~\eqref{eq:feats} needs to be evaluated only once for both the MLP and the SPEN. Therefore, the extra cost of SPEN prediction does not depend on the expense of feature computation. 

We have espoused SPENs' $O(L)$ per-iteration complexity and number of parameters. Strictly-speaking, problems with larger $L$ may require a more expressive global energy network~\eqref{eq:le} with more rows in the measurement matrix $C_1$. This dependence is complex and data-dependent, however. Since the dimension of $C_1$ affects model capacity, a good value depends less on $L$ than the amount of training data. 

Finally, it is difficult to analyze the ability of SPEN prediction to perform accurate global optimization. However, we can establish an upper bound on its performance, by counting the number of times that the energy returned by our optimization is greater than the value of the energy evaluated at the ground truth. Unfortunately, such a search error occurs about 8\% of the time on the Bibtex dataset.

\section{Conclusion and Future Work}
Structured prediction energy networks employ deep architectures to perform representation learning for structured objects, jointly over both $x$ and $y$. This provides straightforward prediction using gradient descent and an expressive framework for the energy function. We hypothesize that more accurate models can be trained from limited data using the energy-based approach, due to superior parsimony and better opportunities for practitioners to inject domain knowledge. Deep networks have transformed our ability to learn hierarchies of features for the inputs to prediction problems. SPENs provide a step towards using deep networks to perform automatic structure learning.

Future work will consider SPENs that are convex with respect to $y$~\citep{amosinput2016}, but not necessarily the model parameters, and training methods that backpropagate through gradient-based prediction~\citep{domke2012generic}.

\section*{Acknowledgements}
This work was supported in part by the Center for Intelligent Information Retrieval and in part by NSF grant \#CNS-0958392. Any opinions, findings and conclusions or recommendations expressed in this material are those of the authors and do not necessarily reflect those of the sponsor. Thanks to Luke Vilnis for helpful advice. 
{\small
\bibliography{sources.bib}
\bibliographystyle{icml2016}
}
\newpage
\appendix
\section{Appendix}

\subsection{Analysis of Convergence Behavior}
\label{app:convergence}

\begin{figure}[!htb]
\centering
\begin{minipage}[b]{\linewidth}
\centering
\includegraphics[width=0.5\columnwidth]{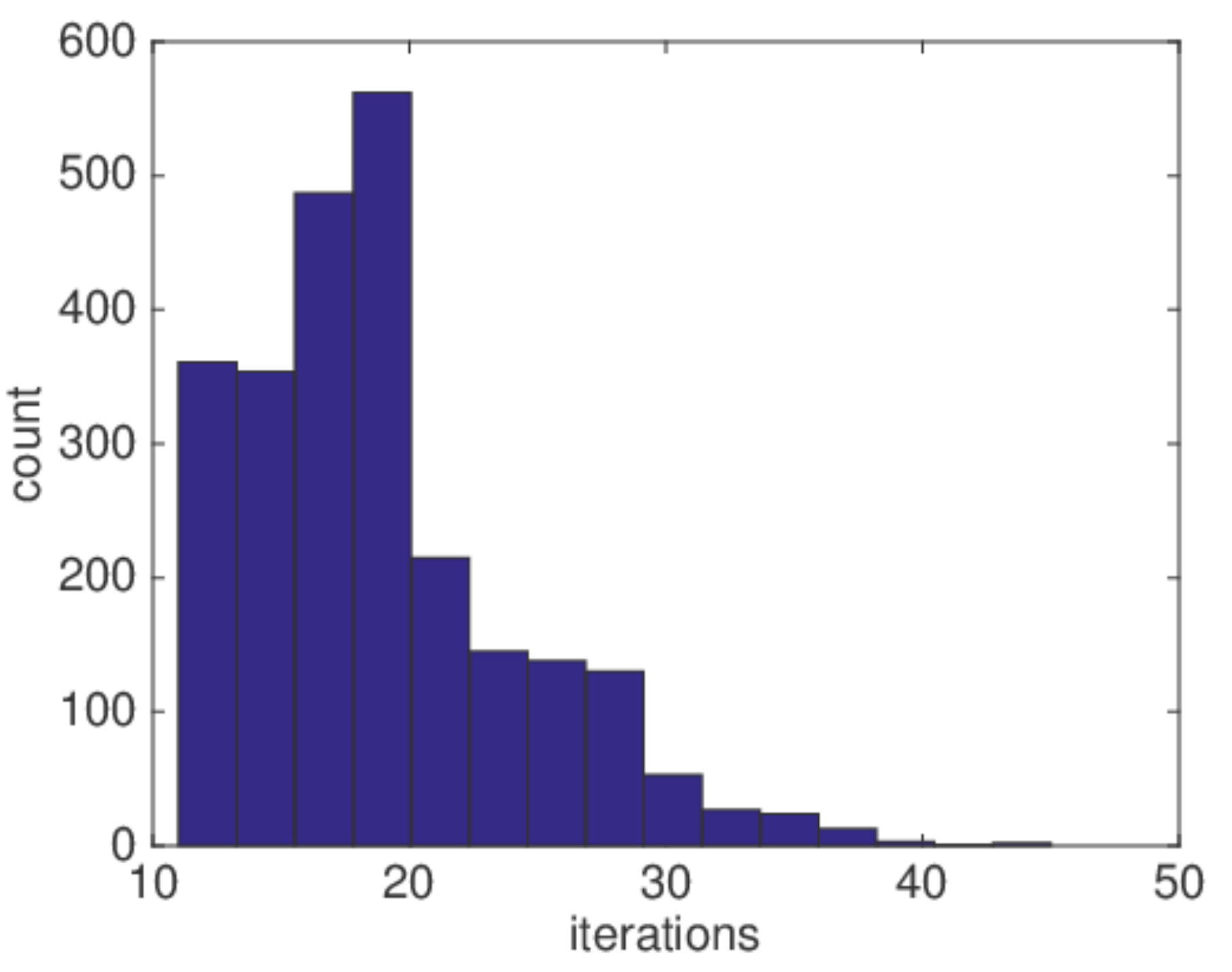}
\caption{Histogram of \#required iters for each example in a large batch to converge. Unnecessary computation is performed on already-converged examples while waiting for the final ones to converge. }
\label{fig:minipage1}
\end{minipage}
\quad
\begin{minipage}[b]{\linewidth}
\centering
\includegraphics[width=0.5\columnwidth]{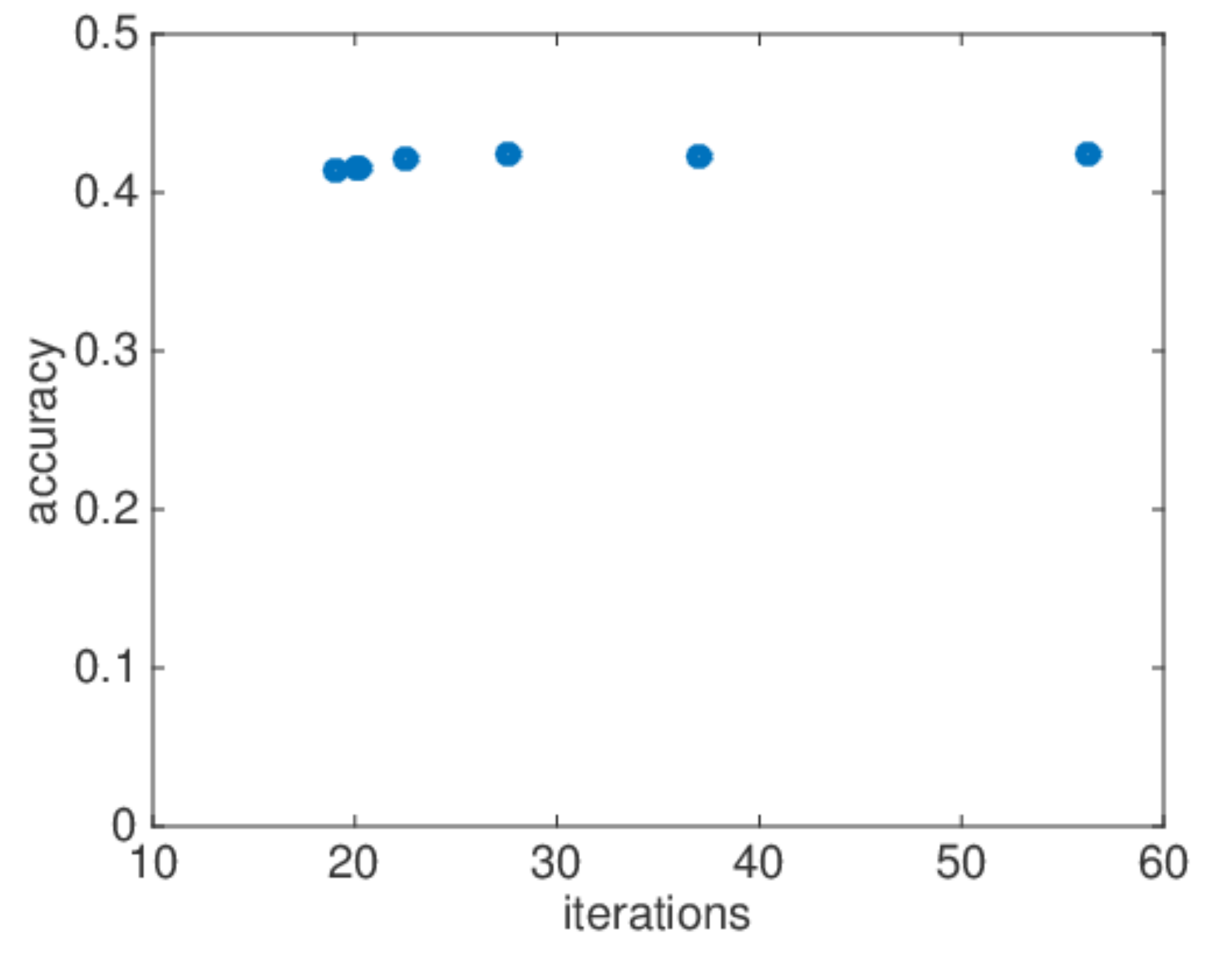}
\caption{Accuracy vs. \# required iters for varying convergence percentage. Optimization on a batch is terminated if X\% of the examples in the batch have converged. This provides little decrease in accuracy, while providing an impressive speedup. We hypothesize that the many of the slow-converging examples were ones for which prediction was going to be incorrect anyway, so it is ok to terminate these early. }
\label{fig:minipage2}
\end{minipage}
\quad
\begin{minipage}[b]{\linewidth}
\centering
\includegraphics[width=0.5\columnwidth]{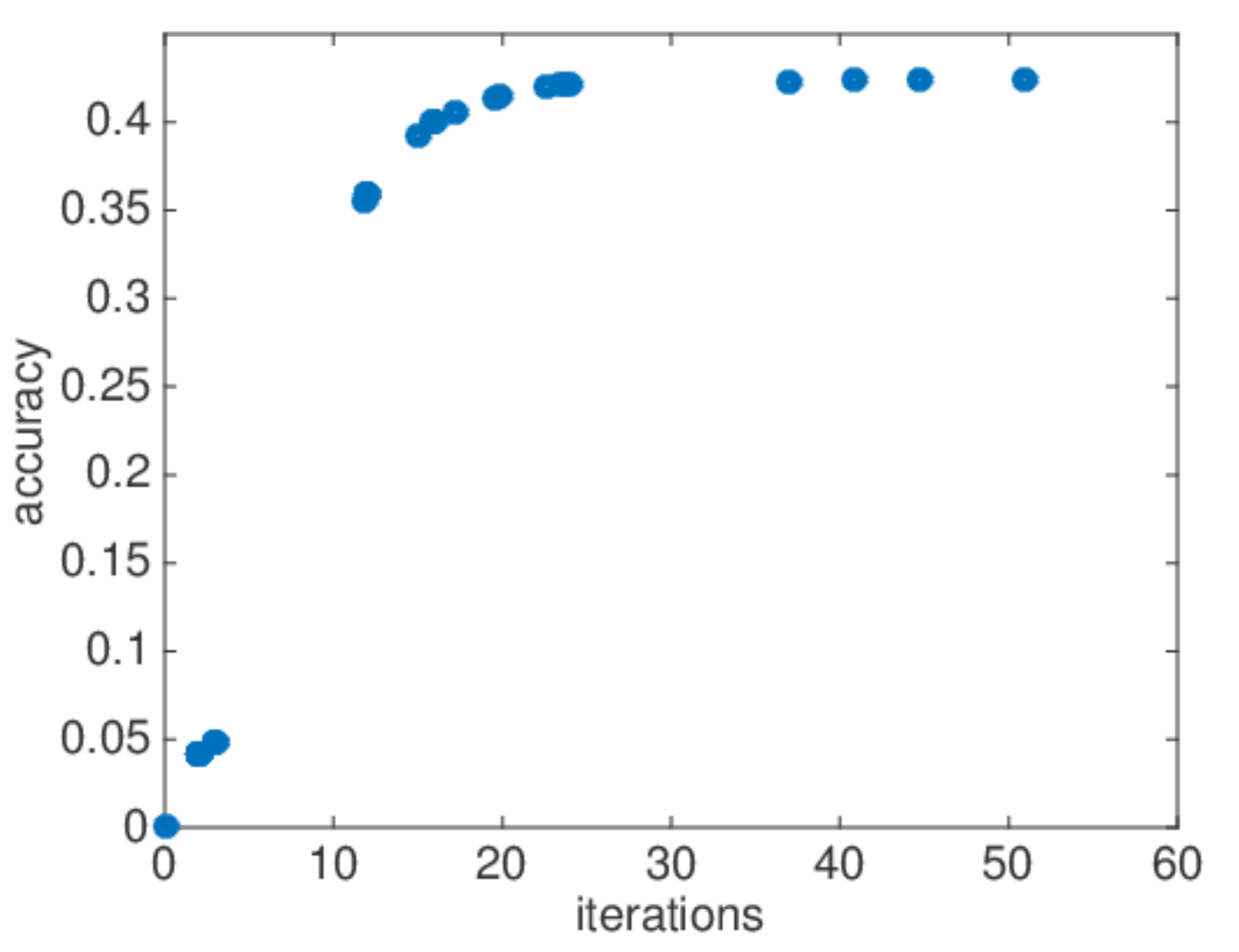}
\caption{Accuracy vs. \# required iters curve for varying convergence tolerance. By using a looseer convergence tolerance, we can sacrifice accuracy for speed. }
\label{fig:minipage3}
\end{minipage}
\end{figure}

\FloatBarrier

\subsection{SPEN Architecture for Multi-Label Classification}
\label{app:architecture-figure}
\includegraphics[width=\columnwidth]{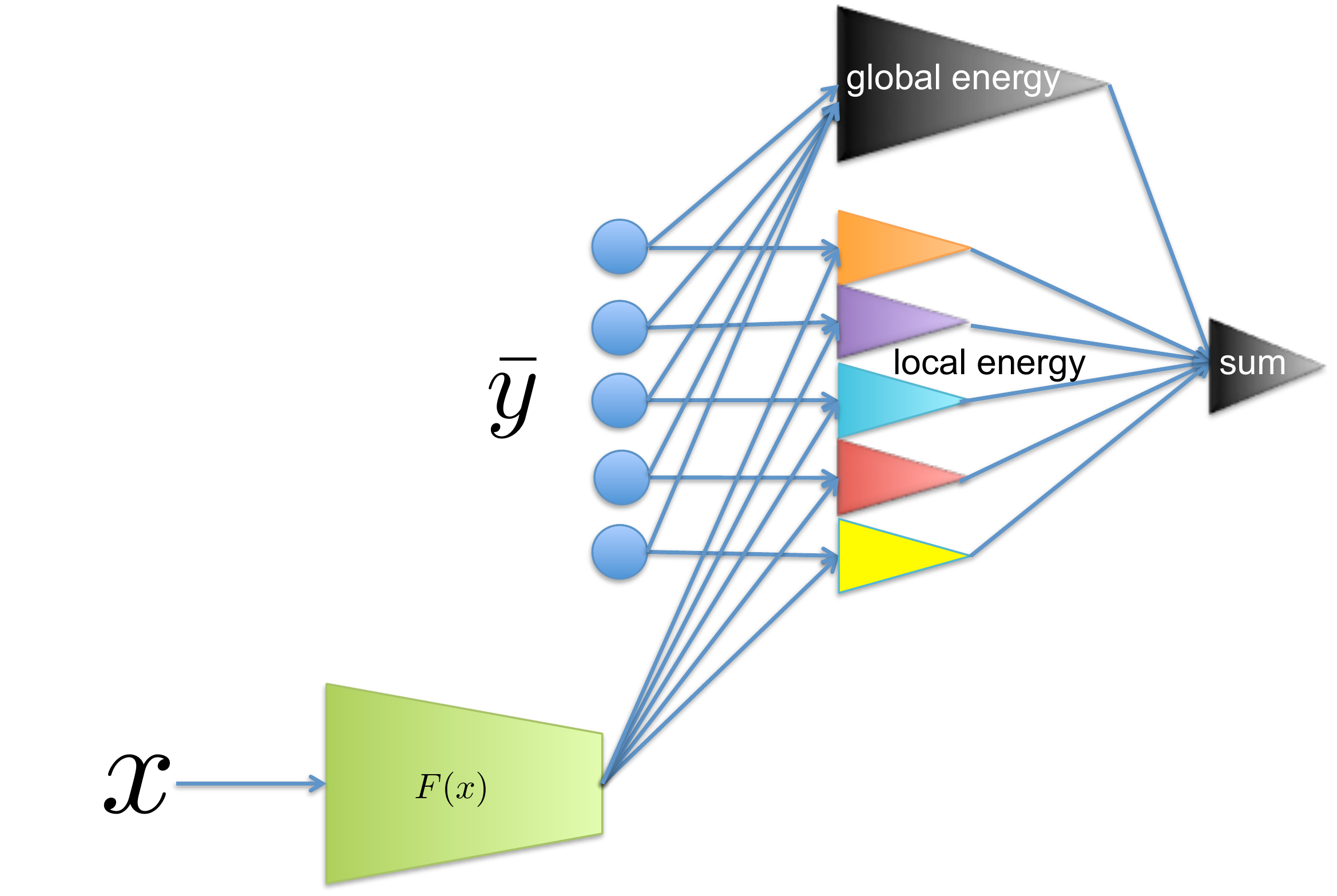}

\subsection{Deep Mean Field Predictor}
\label{sec:dmf-pred}

Consider a fully-connected pairwise CRF for multi-label prediction. We have:
\begin{equation}
P(y | x) \propto \exp\left( \sum_{i,j} B^{(x)}_{ij}(y_i,y_j)  + \sum_i U^{(x)}_i(y_i)\right) \label{eq:exp-family}
\end{equation}
Here, the dependence of the pairwise and unary potential functions $B$ and $U$ on $x$ is arbitrary and we leave this dependence implicit going forward.  There are 4 possible values for $B_{ij}$, dependin\
g on the values of $y_i$ and $y_j$. Similarly, there are two possible values for $U_i$.
Suppose that $y$ is represented as a vector in $\{0,1\}^L$, then we can re-write~\eqref{eq:exp-family} as
\begin{eqnarray}
& P(y | x) \propto \exp( y^\top A_1 y + (1 - y)^\top A_2 y\\
&+ (1 - y)^\top A_3 (1 - y) + C_1^\top y +  C_2^\top (1 - y) )   \label{eq:exp-family}
\end{eqnarray}

Here, $A_1$,$A_2$, and $A_3$ are matrices and $C_1$ and $C_2$ are vectors. Collecting terms, we obtain a matrix $A$ and vector $C$ such that
$$P(y | x) \propto \exp\left( y^\top A y +  C^\top y  + \text{constant}\right)$$
ie
\begin{eqnarray}
P(y | x) \propto \exp\left( y^\top A y +  C^\top y \right)  \label{eq:exp-family2}
\end{eqnarray}

We seek to perform mean-field variational inference in this CRF. Let $\bar{y}^t \in [0,1]^L$ be the estimate for the marginals of $y$ at timestep $t$. Define $\bar{y}^t_{i,0}$ to be a vector that is equa\
l to $\bar{y}$ everywhere but coordinate $i$, where it is $0$ (ie we're conditioning the value of the $i$th label to be 0). Similarly, define $\bar{y}^t_{i,1}$
The mean-field updates set
\begin{equation}
\bar{y}^{y+1}= \frac{\exp(e_i^1)}{\exp(e_i^1) + \exp(e_i^0)} = \text{Sigmoid}(e_i^1- e_i^0),
\end{equation}
where:
$$e_i^1 = (\bar{y}^t_{i,1})^\top A \bar{y}^t_{i,1} +  C^\top \bar{y}^t_{i,1} $$
and
$$e_i^0 = (\bar{y}^t_{i,0})^\top A \bar{y}^t_{i,0} +  C^\top \bar{y}^t_{i,0}. $$

Define $s_i = \sum_{j \neq i} A_{i,j} \bar{y}^t$. Many terms in $e_i^1 - e_i^0$ cancel. We're left with
$$e_i^1 - e_i^0 = s_i + C_i.$$
A vectorized form of the mean-field updates is presented in Algorithm~\ref{alg:mf}.
\begin{algorithm}[tb]
   \caption{Vectorized Mean-Field Inference for Fully-Connected Pairwise CRF for Multi-Label Classification}
   \label{alg:mf}
\begin{algorithmic}
   \STATE {\bfseries Input:}  $x$, $m$
   \STATE $A,C \leftarrow \text{GetPotentials}(x)$
   \STATE Initialize $\bar{y}$ uniformly as $[0.5]^L$
   \STATE $D \leftarrow \text{diag}(A)$
   \FOR{$t=1$ {\bfseries to} $m$}
        \STATE $E \leftarrow A \bar{y} -  D + C$
        \STATE $\bar{y} \leftarrow \text{Sigmoid}(E)$
   \ENDFOR
\end{algorithmic}
\end{algorithm}

\subsection{Details for Improving Efficiency and Accuracy of SPENs}
\label{sec:details}
Various tricks of the trade from the deep learning literature, such as momentum, can be applied to improve the prediction-time optimization performance of our entropic mirror descent approach described in Section~\ref{sec:dmfn}, which are particularly important because $E_x(\ybar)$ is generally non-convex. 

We perform inference in minibatches in parallel on GPUs. 


When `soft' predictions are useful, it can be useful to augment $E_x(\ybar)$ with an extra term for the entropy of $\ybar$. This can be handled at essentially no computational cost, by simply normalizing the iterates in entropic mirror descent at a certain `temperature.' This is only done at test time, not in the inner loop of learning. 

Typically,  backpropagation computes the gradient of output with respect to the input and also computes the gradient of the output with respect to any parameters of the network. For us, however, we only care about gradients with respect to the inputs $\ybar$ during inference. Therefore, we can obtain a considerable speedup by avoiding  computation of  the parameter gradients.

We train the local energy network first, using a local label-wise prediction loss. Then, we clamp the parameters of the local energy network and train the global energy network. Finally, we perform an additional pass of training, where all parameters are updated using a small learning rate. 

\subsection{Hyperparameters}
\label{sec:hyper}
For prediction, both at test time and in the inner loop of learning, we ran gradient descent with momentum = 0.95, a learning rate of 0.1, and no learning rate decay. We terminated prediction when either the relative change in the objective was below a tolerance or the $l_\infty$ change between iterates was below an absolute tolerance. 

For training, we used sgd with momentum 0.9 with learning rate and learning rate decay tuned on development data. We use l2 regularization both when pre-training the features and net and during SSVM training, with l2 weights tuned on development data. 

We did not tune the sizes of the hidden layers for the feature network. These were set based on intuition and the size of the data, the number of training examples, etc.

\begin{table}[t]
\begin{center}
\begin{tabular}{| c | c | c | c | c |}
 \hline 
 &\#labels &\#features & \# train  & \% true labels\\
\hline 
Bibtex & 159 & 1836 & 4880 & 2.40\\
\hline
Delicious & 983 & 500 & 12920 & 19.02\\
\hline
Bookmarks & 208 &  2150 & 60000 & 2.03\\
\hline
Yeast & 14 &  103 & 2417 & 30.3\\
\hline
\end{tabular}
\caption{Properties of the datasets. }
\label{tab:datasets}
\end{center}
\end{table}


\begin{figure}[t]
\centering
\includegraphics[width=0.4\columnwidth]{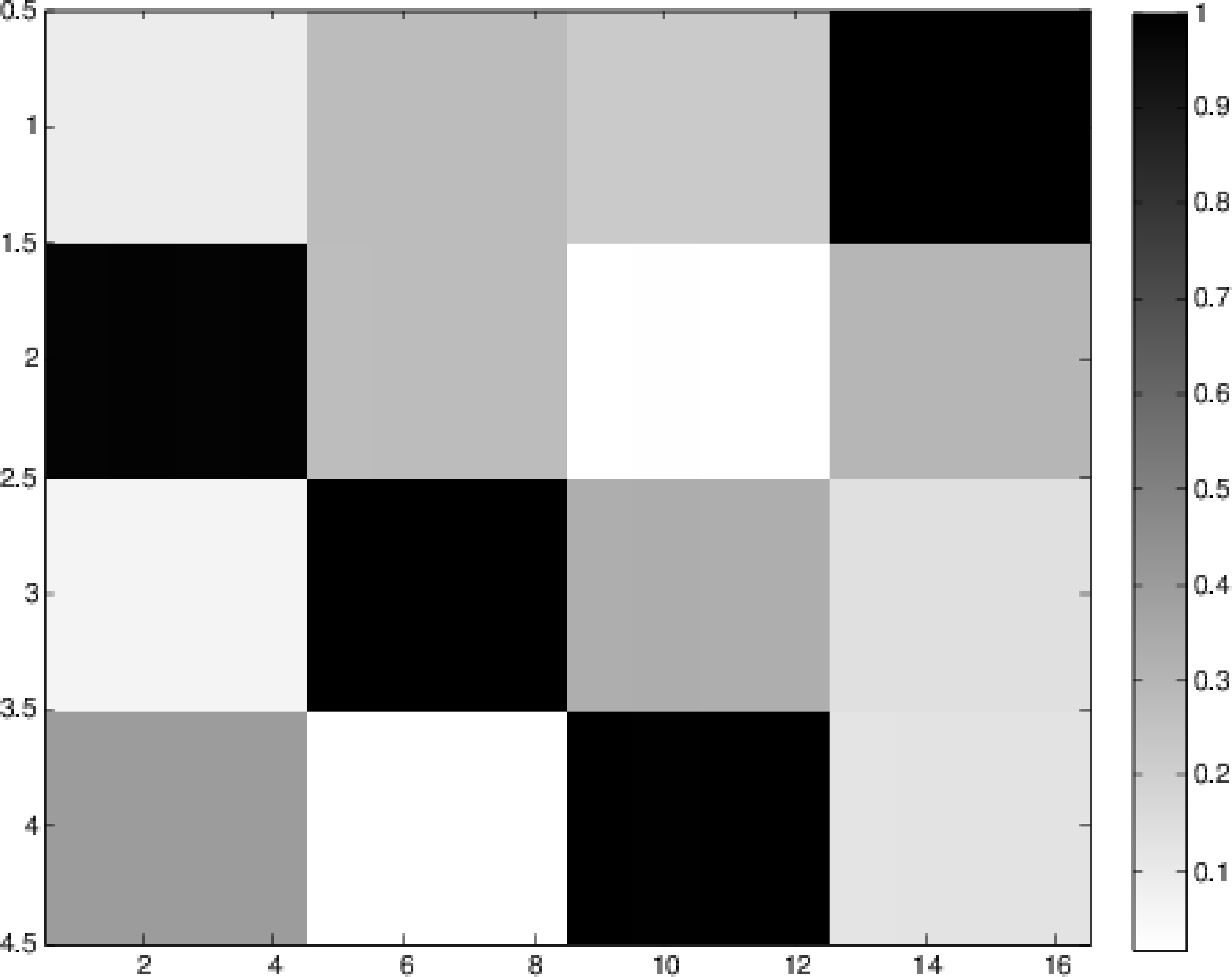}
\caption{Structure learning on synthetic task using 10\% of the data. The measurement matrix still recovers interactions between the labels characteristic of the data generating process}
\label{fig:structure2}
\end{figure}

\end{document}